\documentclass{article}

\usepackage[final]{neurips_style/neurips_2024} % to compile a preprint version

\usepackage[utf8]{inputenc} % allow utf-8 input
\usepackage[T1]{fontenc}    % use 8-bit T1 fonts
\usepackage{hyperref}       % hyperlinks
\usepackage{url}            % simple URL typesetting
\usepackage{booktabs}       % professional-quality tables
\usepackage{amsfonts}       % blackboard math symbols
\usepackage{nicefrac}       % compact symbols for 1/2, etc.
\usepackage{microtype}      % microtypography
\usepackage{xcolor}         % colors
\usepackage{graphicx}       % Required for including graphics
\usepackage{amsmath}        % For math environments
\usepackage{amssymb}        % For math symbols

\usepackage[colorinlistoftodos, textwidth=35mm, shadow]{todonotes}
\usepackage{comment}
\usepackage{hyperref}
\usepackage{subcaption}
\usepackage{amsmath}
\usepackage{amssymb}
\usepackage{algorithm}
\usepackage{algpseudocode}
\usepackage{bbm}
\usepackage{cleveref}

\title{Explaining Concept Drift through the Evolution of Group Counterfactuals}

\author{%
   Ignacy St\k{e}pka\\ % \thanks{Use footnote for providing further information about author (webpage, alternative address)---\emph{not} for acknowledging funding agencies.} \\
   Poznan University of Technology\\
   \And 
   Jerzy Stefanowski \\
   Poznan University of Technology\\
}

\begin{document}

\maketitle

\begin{abstract}
    % Machine learning models deployed in real-world scenarios often face concept drift, where the underlying data distribution changes over time, potentially degrading model performance. While Explainable AI aims to make models transparent mostly for static data sets, explaining the drift itself remains a challenge. This paper proposes a novel approach to explain concept drift by analyzing the evolution of group-based counterfactual explanations (GCEs).  By quantifying shifts in GCE cluster centroids and their associated counterfactual action vectors, we provide interpretation on how the model's decision-making behavior has been altered by the drift and provide more information on the characteristics of local drifts.  We also show how they can be complementary combined with other information on changes coming from the data level and model predictions offering as a result a more comprehensive explanation of types and origins of the drifts. It is illustrated by computational experiments with several case studies.

Machine learning models in dynamic environments often suffer from concept drift, where changes in the data distribution degrade performance. While detecting this drift is a well-studied topic, explaining how and why the model's decision-making logic changes still remains a significant challenge. In this paper, we introduce a novel methodology to explain concept drift by analyzing the temporal evolution of group-based counterfactual explanations (GCEs). Our approach tracks shifts in the GCEs' cluster centroids and their associated counterfactual action vectors before and after a drift. These evolving GCEs act as an interpretable proxy, revealing structural changes in the model's decision boundary and its underlying rationale. We operationalize this analysis within a three-layer framework that synergistically combines insights from the data layer (distributional shifts), the model layer (prediction disagreement), and our proposed explanation layer. We show that such holistic view allows for a more comprehensive diagnosis of drift, making it possible to distinguish between different root causes, such as a spatial data shift versus a re-labeling of concepts.

%\textcolor{red}{We maybe also are intertwining these interpretations with data and model-layer analysis} %This method aims to offer actionable insights into drift dynamics, moving beyond traditional model performance metrics or abstract statistical measures.
%We leverage a 3-layer framework (Data, Model, Explanation) to track changes in GCEs generated before and after a detected drift.
\end{abstract}

\section{Introduction}
\label{introduction}

The development of machine learning (ML) in decision-making and critical applications calls for transparency and trustworthiness. Explainable AI (XAI) has emerged to address this by providing insights into the behavior of complex black-box ML models \cite{molnar2022}. A variety of methods to explain these models have been proposed \cite{bodria2023xai}, but almost all are designed for fully available static data sets. However, many real-world systems operate on data streams that evolve over time -- a phenomenon referred to as concept drift, where the underlying data distribution changes \cite{webb2016characterizing}. This drift can render previously accurate models obsolete.

While most research has focused on detecting drift and adapting models to it \cite{Lu2018}, explaining the nature of the drift itself from a model's perspective has been less explored. Understanding the causes of drift and its impact on the predictions of a model remains largely undeveloped, despite calls from researchers \cite{hinderModelBasedExplanations,webb2016characterizing}. This includes distinguishing local from global drift, locating where it occurs, and analyzing its internal characteristics. Such explanations could help identify whether the class distribution undergoes a temporal decomposition into sub-concepts that may move, split, merge, or even disappear \cite{brzezinski2021impact}, which proves valuable for debugging models and designing better responses to drift.

\begin{figure}[tb]
    \centering
    \includegraphics[width=0.9\linewidth]{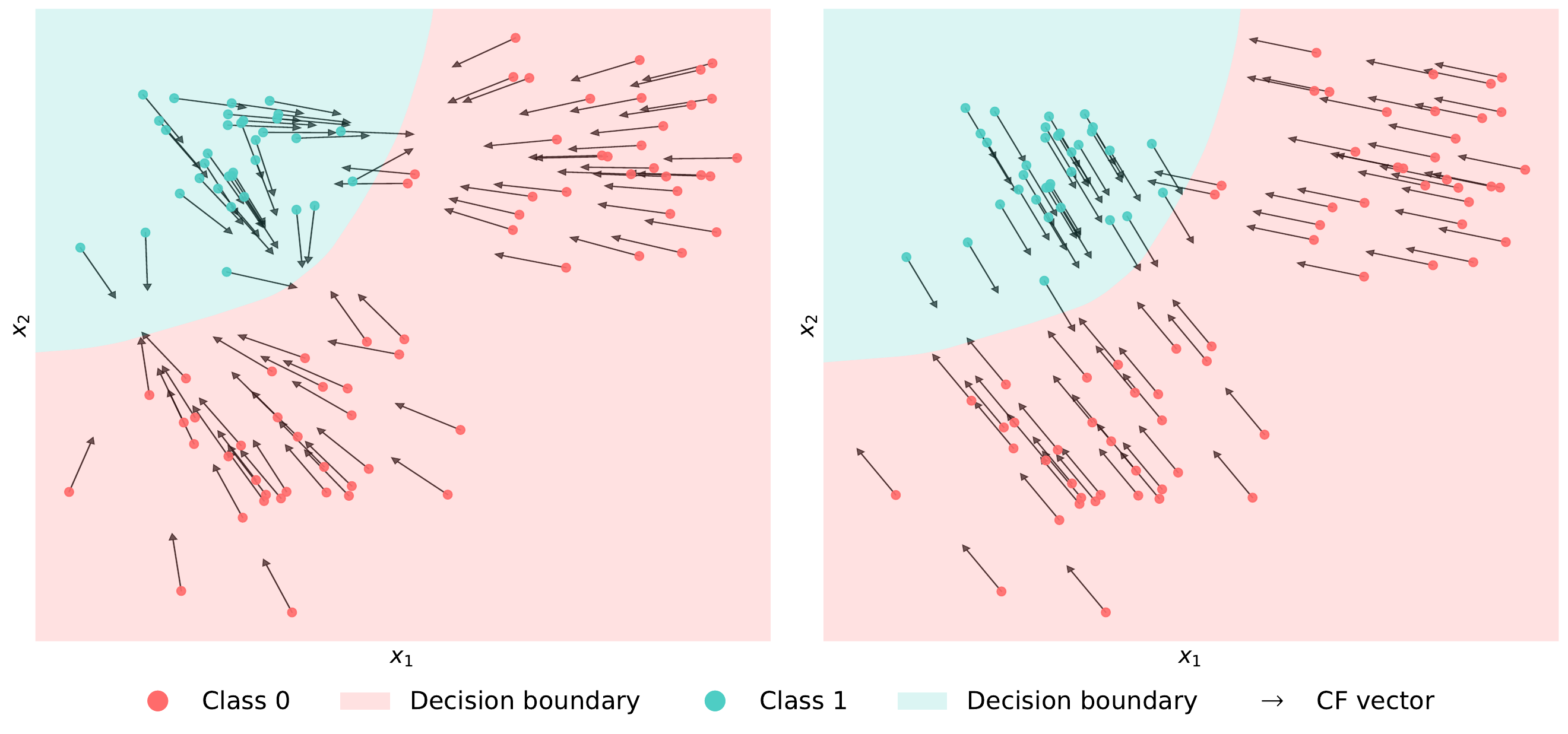}
    \caption{A visualization of the difference between per-sample counterfactual explanations (left) and group counterfactual explanations (right). Note that for GCEs, all CEs belonging to the same group share the same CFAV.}
    \label{fig:rw:cfvsgce}
\end{figure}

Among XAI methods, counterfactual explanations (CEs) are particularly valuable, offering actionable ``what-if'' scenarios that show how changes to input features can alter model predictions \cite{wachter}. While most research concerns individual CEs \cite{Verma2020Counterfactual,guidotti2024counterfactual}, a newer line of work focuses on group counterfactual explanations (GCEs), which extend per-sample CEs to groups of similar instances sharing an explanation vector \cite{glance,Ley2023,kanamori22a,Wielopolski2024,Warren2023,beyerer2021,Chowdhury_2022}. 
In Fig. \ref{fig:rw:cfvsgce},  we provide an intuitive explanation of how GCEs differ from CEs. GCEs provide a more global, yet still interpretable, view of the model's behavior. 
They are well-suited for studying local drifts as they can identify general patterns in data subsets, making them ideal for analyzing the changes in sub-concepts and decision boundaries discussed above. 
To the best of our knowledge, GCEs have not been studied in the context of understanding concept drifts.

Therefore, in this paper, we introduce a novel methodology for explaining concept drift by analyzing the evolution of GCEs. 
We study the evolution of cluster centroids and their associated counterfactual action vectors\footnote{They indicate changes of attribute values needed to obtain a counterfactual. This vector is shared by all examples inside a group.} before and after a drift, showing that they serve as a direct and interpretable proxy for the drift's impact on the model's behavior. To structure this analysis, inspired by the position paper \cite{threeleyers}, we combine GCEs with information from the data layer (distributional changes) and the model layer (prediction changes). We argue that this three-layer approach provides a more comprehensive explanation of drift's type and origin than any single layer can offer alone. 
For our GCE-based analysis of streams with sub-concepts, we adapt the lightweight and extensible GLANCE algorithm \cite{glance}.

Through experimental case studies on different types of drift, we demonstrate that our proposed approach provides a multi-faceted view of the root causes of concept drift. By discussing the benefits and limitations of these cases, we aim to encourage further study into using group counterfactual explanations to analyze and explain concept drifts.

The paper is organized as follows. In Section \ref{related_work} we briefly discuss the context of related works on XAI and in particular group counterfactual explanations. Then, in Section \ref{methodology} we introduce a new method for data and concept shift analysis, which concerns the temporal evaluation of group counterfactuals and the three-layer framework. Next, we present a few case studies in Section \ref{cases}, followed by a discussion (Section \ref{conclusion}) and a note on future directions (Section \ref{sec:limitations}).

\section{Related Work}
\label{related_work}

In data streams, an example $\mathbf{x}$ at time point $t$ is drawn from a joint probability distribution $p_t(x,y)$. Concept drift occurs if for two distinct time points $t$ and $t + \Delta$, $p_t(x,y) \neq p_{t + \Delta}(x,y)$. 
Researchers often focus on a \textit{real drift}, characterized by changes in the posterior probability $p_t(y|x)$. Drifts can be categorized by their rate of change (sudden, gradual, incremental) and their scope (global, local) \cite{webb2016characterizing,brzezinski2021impact,Lu2018}. In this paper, we are particularly interested in drifts that involve the decomposition of classes into sub-concepts \cite{brzezinski2021impact}.

A wide variety of methods exist for detecting drift \cite{Lu2018,informaticsdrift}. Most approaches aim to identify a change by monitoring either the model's predictive performance or the data distribution itself. The latter often involves comparing statistical parameters (e.g., feature means, variances) of the data distribution pre- and post-drift \cite{Lukats2025}. Analysis of the model typically involves tracking predictive metrics, as a significant drop in performance is a common signal for supervised drift detectors \cite{Lu2018}. 
Some studies go further by comparing pre- and post-drift predictions at the instance level to identify regions of disagreement \cite{BobekDomainAdapt,hinderModelBasedExplanations}. However, these methods identify \textit{that} drift has occurred, but generally fall short of explaining \textit{how} the model's underlying logic has changed.

Explaining such model changes is a key challenge for XAI in streaming settings \cite{hammerandhullermeier,hinder2023thingsknowconceptdrift}. XAI for static models has experienced substantial advancements with methods that provide explanations in the form of, e.g., feature attribution, partial dependence plots, rule-based explanations , and counterfactual explanations, for their review see e.g \cite{bodria2023xai,molnar2022}. 
However, adapting them to dynamic data is not straightforward. 
In streaming settings, where models must adapt to non-stationary data, XAI faces additional challenges. Much of the existing work extends static methods to operate online or focuses on local, per-instance explanations.  Some efforts have been made to characterize such changes using feature importance~\cite{Muschalik2022,shapAdwin,chen2022explaining} and partial dependence plots~\cite{Muschalik2023_PDP}, but comprehensive explanations of concept drift remain an open challenge~\cite{hammerandhullermeier,hinder2023thingsknowconceptdrift}.

A more targeted line of research attempts to explain drift using contrastive examples~\cite{hinderModelBasedExplanations}, which provide localized insight and support the drift localization. Artelt et al.~\cite{Artelt2023Contrasting} advance this by applying CFEs to characterize concept drift through changes in distance (e.g., norm, cosine similarity) between counterfactuals. In addition, they propose regularizing future model updates to preserve past counterfactuals. These contributions open a promising direction, yet remain limited in scope.

Our work builds on these foundations by integrating counterfactual explanations into a broader explanatory framework. 
Specifically, we utilize group counterfactual explanations \cite{glance,Ley2023,kanamori22a,Wielopolski2024,Warren2023,beyerer2021,Chowdhury_2022,rawal2020beyond}\footnote{Sometimes also named multi-instance counterfactual explanations \cite{artelt2024two}}, which explain groups of instances in a collective way sharing the same counterfactual vectors among many examples (see Fig.~\ref{fig:rw:cfvsgce}). GCEs reduce the number of counterfactual directions from $N$ (one per instance) to $k$ (one per group, where $k \ll N$), improving interpretability. This summarization enables clearer insight into how segments of the input space would need to change to alter model predictions. 

We note that GCEs are particularly suited for the streaming setting. Once established, they allow for efficient issuing of new counterfactuals via line search along shared group directions, avoiding costly full optimization for each instance. In this work, we adapt the GLANCE~\cite{glance} method for use in data streams. GLANCE forms $k$ clusters per class, each with a corresponding counterfactual action vector (CFAV) and supporting spatial localization.

Our approach diverges from previous work not just by using GCEs, but by analyzing how they \textit{evolve} over time. 
We posit that the temporal changes in GCE cluster centroids and their corresponding counterfactual action vectors constitute a powerful and interpretable signal for explaining drift. 
Where prior work has characterized drift through the lens of individual examples, our method provides a more structured and global view, revealing how the model's logic for entire sub-populations is altered by the drift.

\section{Method}
\label{methodology}

Our approach monitors the incoming data stream, denoted as $D_t$ at time $t$, and the prediction performance of a model $h$ learned from it. Upon detection of a drift, we denote the new data context as $D_{t+1}$. 

In order to explain the signaled drift and analyze its causes, we propose to consider information from the data, model, and explanation layers in a unified framework, which we describe in the following subsections.

\subsection{GCEs evolution as drift explanation}
This layer is central to our proposal, as we leverage GCEs to understand how the model's decision boundaries and rationale shift due to drift. 

\paragraph{Counterfactual Explanations} 
Counterfactual explanations aim to identify a minimal change to an input example $\mathbf{x}$ that leads to a different prediction from a model $h$. Specifically, they seek a counterfactual action vector $\mathbf{c}$ such that the modified input $\mathbf{x} + \mathbf{c}$ results in a different outcome, i.e., $h(\mathbf{x}) \neq h(\mathbf{x} + \mathbf{c})$. Simultaneously, they aim to remain close to the original input according to some distance metric $d(\mathbf{x}, \mathbf{x} + \mathbf{c})$ (inspired by the formulation of Wachter et al. \cite{wachter}), where $\mathbf{c}$ is obtained by solving the following optimization problem:

\begin{equation}
    \mathbf{c} = \arg\min_{\mathbf{c}} d(\mathbf{x}, \mathbf{x} + \mathbf{c}) \quad \text{subject to} \quad h(\mathbf{x}) \neq h(\mathbf{x} + \mathbf{c})
\end{equation}

This objective is often optimized as a single composite loss function \cite{wachter}:

\begin{equation}
\label{eq:cfe:definition}
    \mathcal{L}  = \mathcal{L}_{\text{validity}}(h(\mathbf{x}), 1- h(\mathbf{x} + \mathbf{c})) + \lambda \cdot \mathcal{L}_{\text{proximity}}(\mathbf{x}, \mathbf{x} + \mathbf{c}) 
\end{equation}
where $\mathcal{L}_{\text{validity}}$ is a metric that ensures class predictions for the original example $\mathbf{x}$ and the counterfactual $\mathbf{x}+\mathbf{c}$ are different, $\mathcal{L}_{\text{proximity}}$ ensures that the counterfactual stays as close as possible to the original example in the feature space, and $\lambda$ is a trade-off coefficient. Both $\mathcal{L}_{\text{proximity}}$ and $\mathcal{L}_{\text{validity}}$ can be, but are not limited to, being implemented with $L_1$ or $L_2$ distance. 

Depending on the specific method, in addition to validity and proximity, other properties of counterfactuals are often expected \cite{guidotti2024counterfactual,Verma2020Counterfactual}, such as sparsity of feature changes or plausibility -- ensuring the counterfactual is situated within a high-density region of the original data \cite{poyiadzi2020face,Wielopolski2024}. However, no single method excels across all properties \cite{stepka2024multi}. Thus, in this work, we focus on validity and proximity, as the desirability of other properties is often application-specific and outside the scope of this paper.  

\paragraph{Group Counterfactual Explanations}  
Given a desired number of groups $k$, the goal is to learn $k$ counterfactual action vectors $\mathbf{c}_1, \mathbf{c}_2, \ldots, \mathbf{c}_k \in C$, each associated with a corresponding cluster of input examples $s_1, s_2, \ldots, s_k \in S$. Each CFAV $\mathbf{c}_i$ should be small in magnitude to optimize proximity, and when applied to inputs in its assigned cluster $s_i$, should flip the prediction. The objective is similar to Eq.~\ref{eq:cfe:definition}: minimize average distance while maximizing validity, where validity is the proportion of examples for which the CFAV successfully flips the prediction. The difference is in the loss terms, which depend on $S$ and $C$.

\begin{equation}
\mathcal{L}_{\text{validity}} =  \frac{1}{n} \sum_{i=1}^{k} \sum_{\mathbf{x} \in s_i} \mathbbm{1}\left[h(\mathbf{x}) \neq h(\mathbf{x} + \mathbf{c}_i)\right]
\end{equation}
\begin{equation}
\mathcal{L}_{\text{proximity}} =  \frac{1}{n} \sum_{i=1}^{k} \sum_{\mathbf{x} \in s_i} d(\mathbf{x}, \mathbf{x}+\mathbf{c}_i)
\end{equation}
As mentioned in Sec. \ref{related_work}, various existing algorithms focus on different aspects of GCEs. In this work, we use GLANCE \cite{glance} because its optimization goals align with ours. A summary of GLANCE is provided in Alg.~\ref{alg:glance}. It starts by generating base CEs for all samples $X$ using a chosen CE generation method (e.g., Wachter~\cite{wachter}, FACE~\cite{poyiadzi2020face}). Then, it performs hierarchical clustering using a distance metric that combines feature-space $L_2$ distance and the cosine similarity of CFAVs. It iteratively finds the pair of clusters with the smallest combined distance, merges them by computing weighted means of centroids and CFAVs, and repeats until $k$ clusters are formed.

\begin{algorithm}[htb]
\caption{GLANCE}
\label{alg:glance}
\begin{algorithmic}[1]
\Require Data $X$, base counterfactual explanation method $\text{BCE}(\cdot)$, desired number of clusters $k$
\State \textbf{Step 1: Generate Base Counterfactual Explanations}
\State $C \leftarrow \emptyset$ \Comment{\textit{An empty set of CFAVs}}
\For{each $\mathbf{x} \in X$}
    \State $\mathbf{c} \leftarrow \text{BCE}(\mathbf{x})$ \Comment{\textit{Generate base CE}}
    \State $C \leftarrow C \cup \{\mathbf{c}\}$
\EndFor

\State \textbf{Step 2: Cluster using combined distance metric}
\State $M \gets \mathbf{0}^{|X| \times |X|}$ \Comment{\textit{Initialize the matrix of pairwise distances}}
\State $\mathbf{w} \gets \mathbf{1}^{|X|}$ \Comment{\textit{Initialize the vector of group weights}} 
\Repeat
    \For{$i = 1$ to $|M|$, $j = i+1$ to $|M|$} 
        \State $M_{ij} \leftarrow \|\mathbf{x}_i - \mathbf{x}_j\|_2 + \left(1 - \frac{\mathbf{c}_i \cdot \mathbf{c}_j}{\|\mathbf{c}_i\|_2 \|\mathbf{c}_j\|_2}\right)$
    \EndFor
    \State Find pair $(i,j)$ in $M$ with lowest $d_{\text{clust}}$
    \State Compute new centroid $\bar{\mathbf{x}}_{\text{new}} \gets \frac{1}{w_i+w_j} (w_i\bar{\mathbf{x}}_i + w_j\bar{\mathbf{x}}_j)$
    \State Compute new CFAV $\bar{\mathbf{c}}_{\text{new}} \gets \frac{1}{w_i+w_j} (w_i\bar{\mathbf{c}}_i + w_j\bar{\mathbf{c}}_j)$
    \State Shrink matrix $M$ by 1 row and column
    \State Update $w_{\text{new}} \gets w_i + w_j$
\Until{$k$ clusters exist}
\end{algorithmic}
\end{algorithm}

GLANCE produces $k$ clusters and CFAVs, which are used at test time to generate counterfactuals. For a new instance $\mathbf{x}$, the closest centroid $\bar{\mathbf{x}}_k$ is found via $L_2$ distance, and its associated CFAV is used.

We apply GLANCE and the general GCE framework to analyze concept drift. Specifically, we use pre- and post-drift centroids and their respective CFAVs to characterize the nature of the change. We analyze feature-wise differences in GLANCE clusters before and after the drift to assess locality and examine shifts in cluster positions to check if the decision boundary has moved.

We also compare pre- and post-drift CFAVs to assess changes in the decision boundary. If vectors remain stable, the change is likely minor. Significant differences in vectors suggest a larger shift. We quantify this difference by calculating the cosine similarity between vectors and visualizing it per-group. Additionally, we visualize the feature-wise vector directions to identify which features were most affected and what the change entailed. 

To perform this analysis, we must match clusters pre- and post-drift. We use simple one-to-one matching based on the Euclidean distance between centroids. This method is intuitive for end-users, as it aligns with how they interpret similarity in feature space.
Finally, since $k$ is a hyperparameter, we use the x-means \cite{xmeans2000} algorithm as a warmup to estimate $k$. This gives rise to our assumption about the data distribution used in this paper: that it can be decomposed into disjoint, mean-centered subgroups. 

\paragraph{Modifications to GLANCE} The following changes are based on preliminary experiments. While not central to this paper, we note them here for transparency and reproducibility.
Originally, GLANCE uses Wachter~\cite{wachter} for CE generation. However, we found it unstable with different random seeds. Specifically, initialization changes led to large differences in base CEs and consequently in the final groups. To address this, we use FACE~\cite{poyiadzi2020face}, a lightweight method that optimizes for validity, proximity, and plausibility. The last property is crucial for generating stable counterfactuals that are robust to noise in the data or the model \cite{stepka2025robust}. Without this robustness, the proposed drift explanations would be unreliable.

Furthermore, we observed that clusters near the decision boundary could be merged even when they belong to different classes, a finding that we confirmed experimentally. To avoid this, we run GLANCE separately for each predicted class, producing $k$ groups per class. This improved CFAV and centroid stability and increased validity in preliminary experiments.

\subsection{Analyzing Changes in Data Distributions}
To compare data distributions $D_t$ (pre-drift) and $D_{t+1}$ (post-drift), we follow established unsupervised drift analysis techniques of monitoring and comparing the means of individual features, which is a common strategy in drift detection \cite{Lukats2025}. Shifts in feature means can indicate which features are most responsible for the drift and help quantify the overall magnitude of change. We track the means on a per-class basis to allow for a more comprehensive overview and to understand whether the shift is in $P(X)$ or $P(X|Y)$.

Next, if the location of a GCE cluster changes significantly, we can complement this information by determining whether this shift is driven by a change in $P(X)$ or in $P(Y|X)$. This demonstrates how GCEs, combined with data analysis, can help to distinguish whether the change is due to a data distribution shift (real drift) or a concept shift (virtual drift).

\subsection{Online Adaptation and Disagreement}
Let $h_t$ be the model trained on data up to time $t$. Upon detection of drift, $h_t$ is adapted using recent data (e.g., from a sliding window), resulting in a new model $h_{t+1}$. The central assumption is that concept drift will cause a change in the function learned by the model. To assess this, we focus on the change in output behavior and compare $h_t$ and $h_{t+1}$ using the mean absolute error of their predictions. Let $X_{\text{eval}} = D_t \cup D_{t+1}$, then:
\begin{equation}
d_{\text{mae}}(h_t, h_{t+1}, X) = \frac{1}{|X|} \sum_{\mathbf{x} \in X} \vert h_t(\mathbf{x}) - h_{t+1}(\mathbf{x}) \vert
\end{equation}
which we call the \textit{global disagreement coefficient}. In visualizations, we also disaggregate this global notion of model change into local disagreement coefficients, localized around the GCE-generated centroids.  

This model-layer analysis is especially useful for identifying the global magnitude of model change and locating where in the input space disagreement occurs. This localizes regions of maximum model instability, offering direct clues about the areas most affected by concept drift.

\subsection{Synergy of the Three Layers}
We hypothesize that none of the layers -- data, model, or explanation -- is sufficient on its own for a complete drift interpretation. Rather, their interaction provides a more comprehensive and meaningful insight, as postulated in \cite{threeleyers}. We formulate the following expectations for the role of each layer:
\begin{itemize}
    \item The \textbf{Data layer} identifies whether the input distribution $P(X)$ has changed and which features or regions were involved.
    \item The \textbf{Model layer} measures change in the decision function and helps localize regions of significant model disagreement pre- and post-drift.
    \item The \textbf{GCE Explanation layer} shows whether and where the decision boundary and the feature-based counterfactual logic have shifted.
\end{itemize}
We will discuss the synergies of these explanations and the specific role of each in more detail in the experimental cases that follow.

\section{Analysis of Drift Scenarios}
\label{cases}

To validate our proposed three-layer analytical framework, we now demonstrate its efficacy across several distinct concept drift scenarios. 
We employ synthetic datasets where the ground-truth nature of the drift is known \textit{a priori}, enabling a clear evaluation of the contribution of each layer. 
The data is generated from multiple Gaussian distributions that form discrete sub-concepts within each class, which are then subjected to controlled transformations to simulate various types of drift. 
The following case studies illustrate how the synergy between data-level, model-level, and explanation-level insights provides a comprehensive understanding of the drift that is unattainable by any single layer alone.

\subsection{Case 1: Data Shift}

\begin{figure}[htb]
    \centering

    \begin{subfigure}[b]{0.43\textwidth}
        \includegraphics[width=\textwidth]{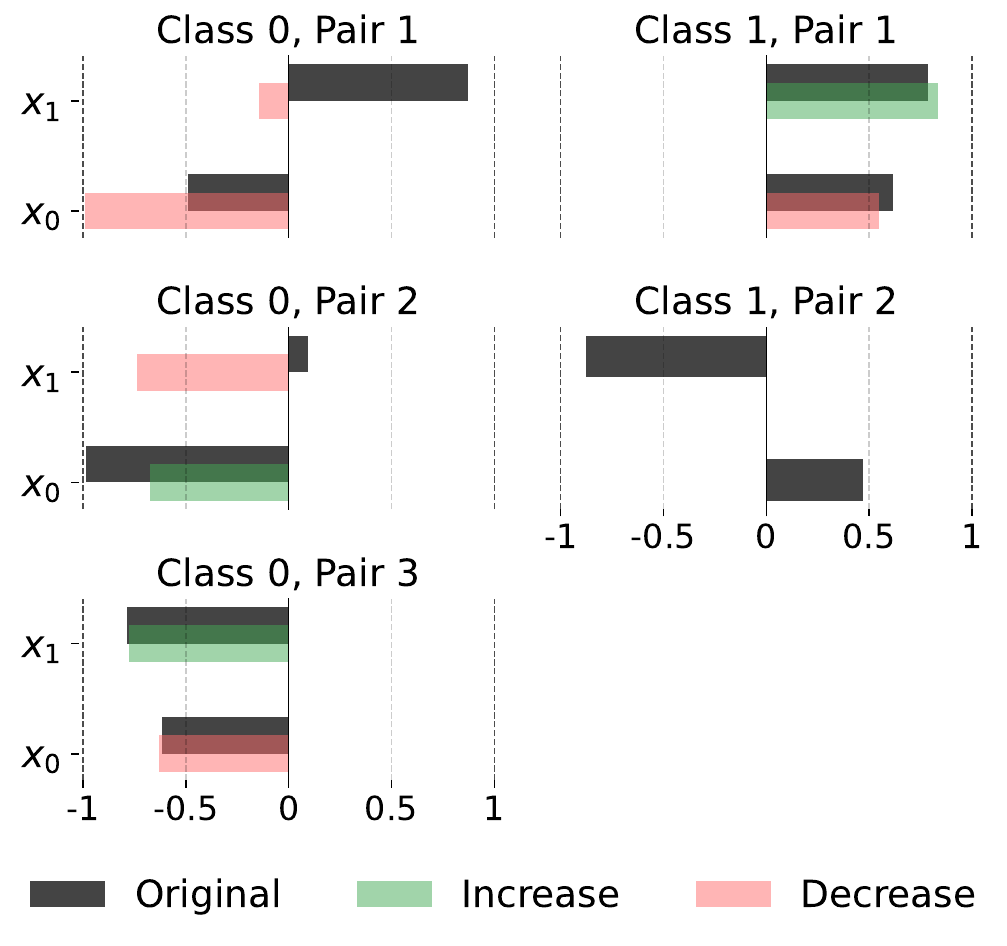}
        \caption{CFAVs}
    \end{subfigure}
    \hfill
    \begin{subfigure}[b]{0.42\textwidth}
        \includegraphics[width=\textwidth]{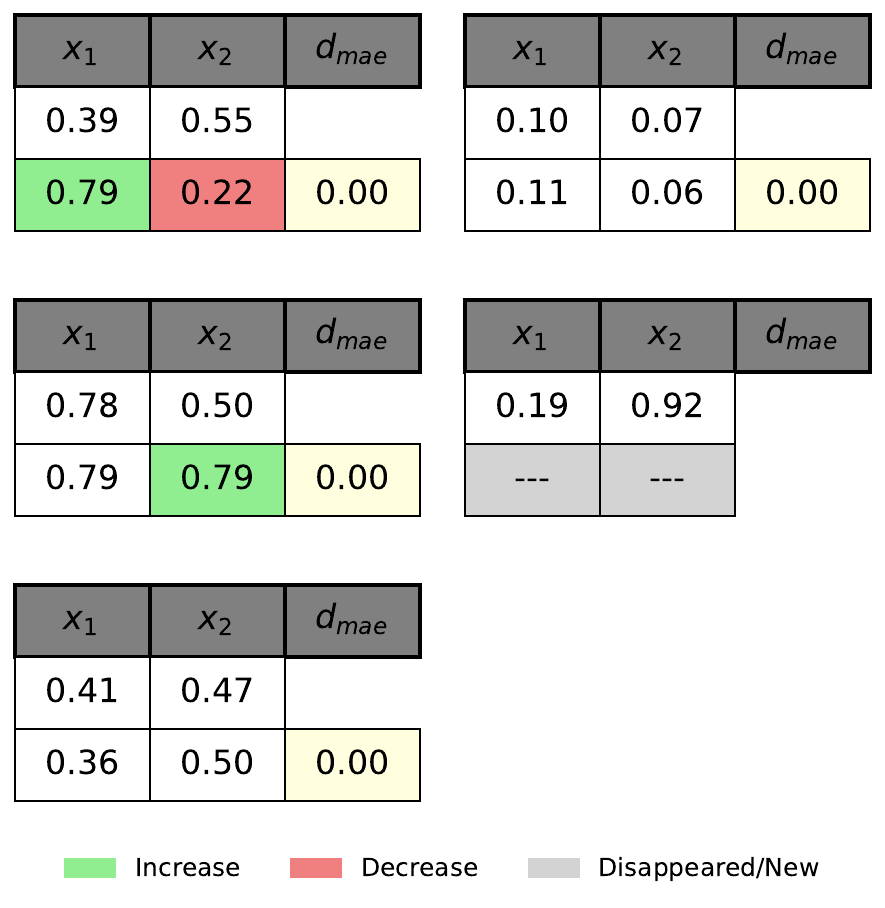}
        \caption{Centroids}
    \end{subfigure}
    \hspace{2cm}
    \hfill
    \begin{subfigure}[b]{0.37\textwidth}
        \includegraphics[width=\textwidth]{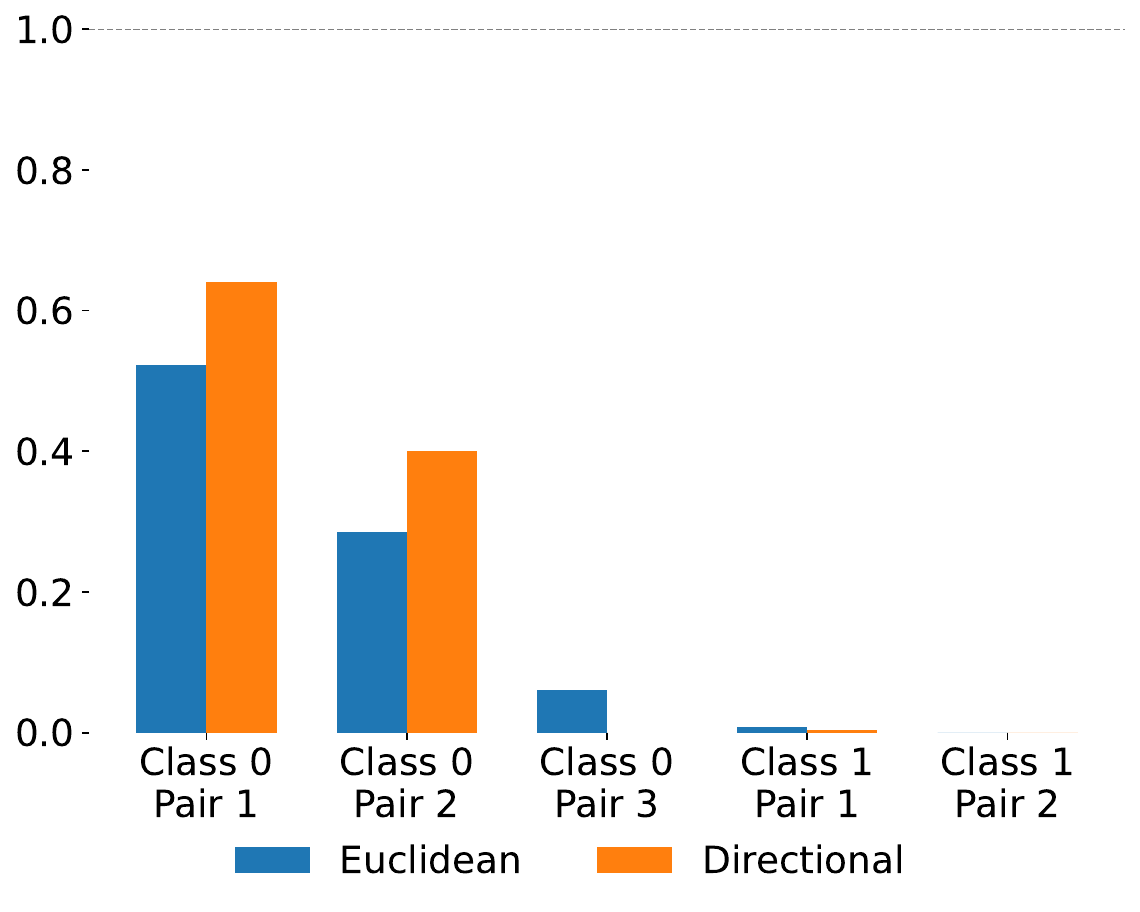}
        \caption{Summary}
    \end{subfigure}
    \hfill
    \begin{subfigure}[b]{0.59\textwidth}
        \includegraphics[width=\textwidth]{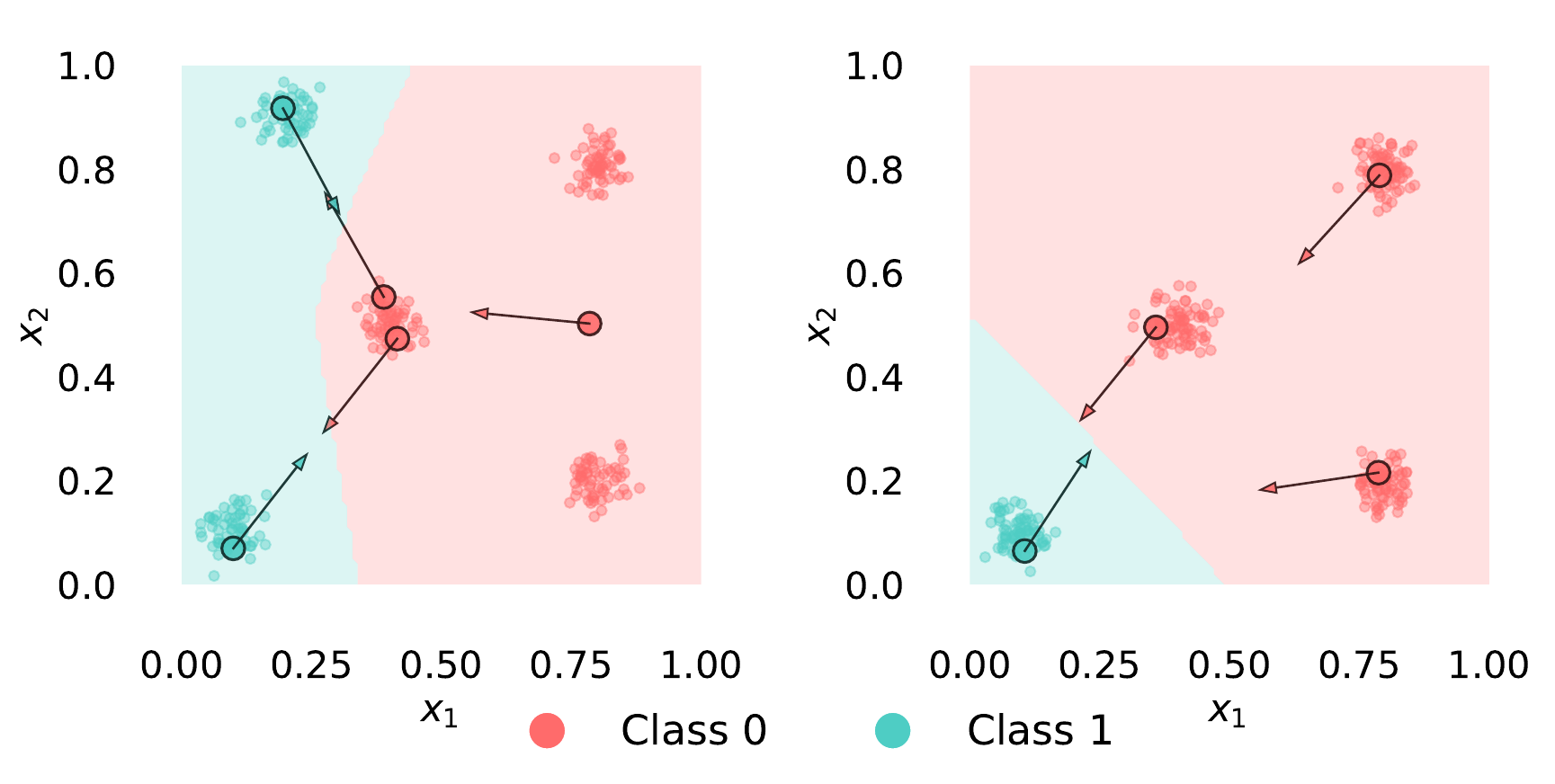}
        \caption{Feature-space visualization}
    \end{subfigure}
    \caption{Analysis of a data shift where one sub-concept disappears (see Fig. \ref{fig:case1}(d), bottom left). \textbf{(a)} Per-feature changes in CFAVs. \textbf{(b)} Pre- and post-drift GCE centroid locations and local model disagreement ($d_{mae}$). \textbf{(c)} Summary of Euclidean (CFAVs) and directional changes (centroids). \textbf{(d)} Visualization of data and GCEs pre-drift (left) and post-drift (right). \textbf{Global disagreement}: $d_{mae} \approx 0.0$. \textbf{Per-class feature means}: \textit{before} -- (\textbf{0:} $x_1$=0.67, $x_2$=0.5, \textbf{1:} $x_1$=0.15,$x_2$=0.5), \textit{after} -- (\textbf{0:} $x_1$=0.67,$x_2$=0.5, \textbf{1:} $x_1$=0.1,$x_2$=0.1).}
    \label{fig:case1}
\end{figure}

We first consider a drift characterized by a change in the data distribution $P(X)$, where a sub-concept belonging to Class 1 vanishes (Fig. \ref{fig:case1}).

\textbf{Data:} An analysis of per-class feature means reveals a shift for Class 1 (the mean of $x_2$ changes from 0.5 to 0.1), while the statistics for Class 0 remain stable. Although this indicates a change has occurred within Class 1, this aggregated view lacks the spatial or structural information necessary to understand its nature.

\textbf{Model:} The global disagreement coefficient remains at 0.0, indicating that the retrained model, $h_{t+1}$, produces predictions nearly identical to $h_t$ on the available data. This layer of analysis is therefore unable to detect the drift, likely because the vanished data cluster had a negligible impact on the final decision boundary in regions with data support.

\textbf{Explanation:} In contrast, the GCE layer provides critical insight. 
It identifies the vanished sub-concept, noting that the cluster corresponding to \textit{Class 1, Pair 2} has disappeared post-drift (Fig. \ref{fig:case1}(b)). 
While a 2D feature-space plot confirms this graphically, our analysis relies on the tabular and vector-based outputs, which scale to higher dimensions. 
More subtly, even though the model's predictions on data did not change ($d_{mae} \approx 0.0$), the CFAVs for the remaining clusters show moderate Euclidean and directional changes (Fig. \ref{fig:case1}(a,c)). 
This demonstrates that the model's underlying counterfactual logic \textit{has} shifted in response to the drift, a nuance entirely missed by the other layers. 
Thus, the GCE-based analysis was capable of providing a clear, localized, and interpretable account of the drift, moving beyond descriptive statistics to identify precisely which sub-concept vanished and how its disappearance impacted the classifier's rationale.

\subsection{Case 2: Real Concept Drift}

\begin{figure}[h!]
    \centering
    \begin{subfigure}[b]{0.4\textwidth}
        \includegraphics[width=\textwidth]{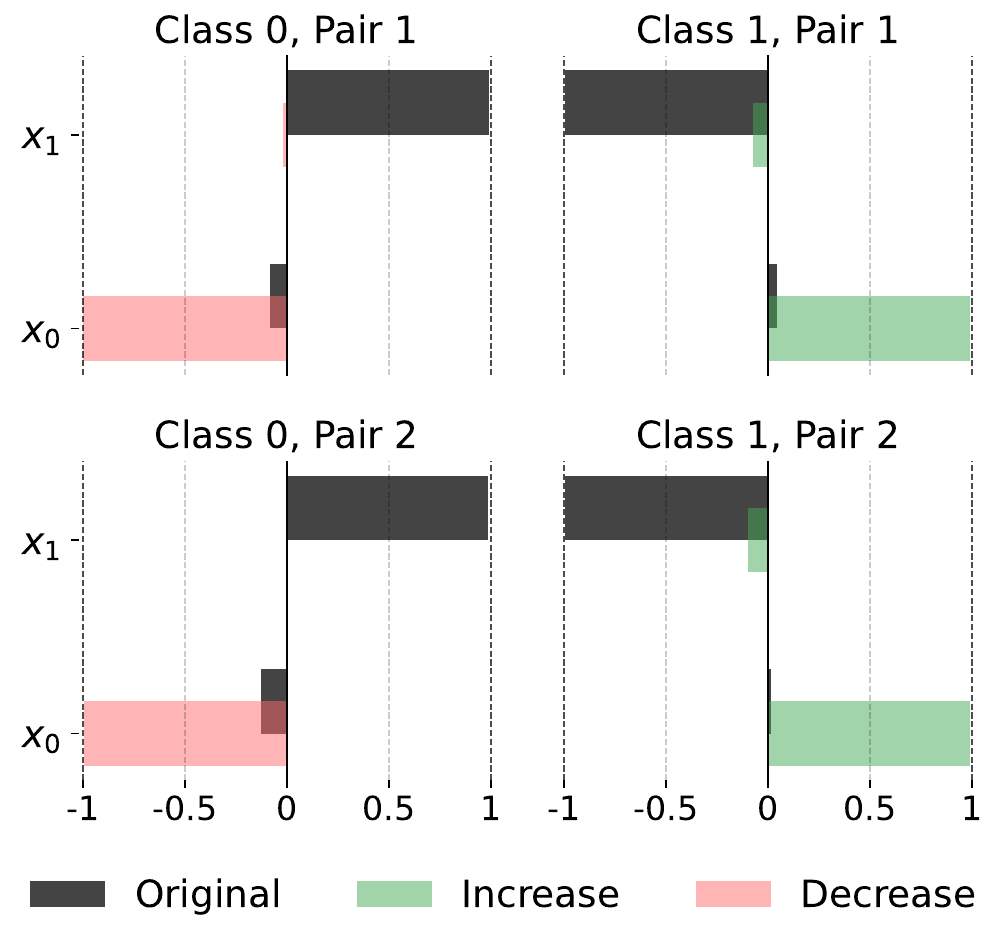}
        \caption{CFAVs}
    \end{subfigure}
    \hfill
    \begin{subfigure}[b]{0.5\textwidth}
        \includegraphics[width=\textwidth]{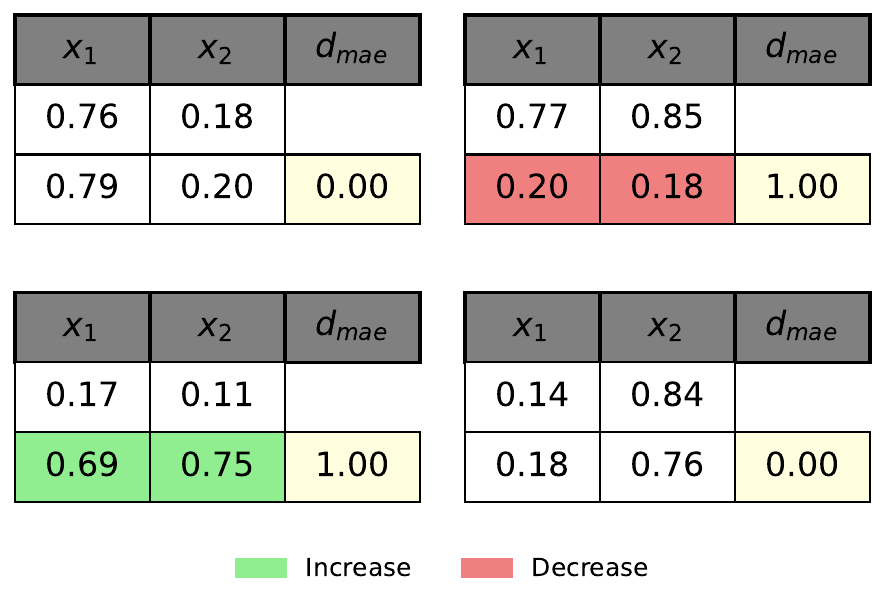}
        \caption{Centroids}
    \end{subfigure}
    \hfill
    \begin{subfigure}[b]{0.4\textwidth}
        \includegraphics[width=\textwidth]{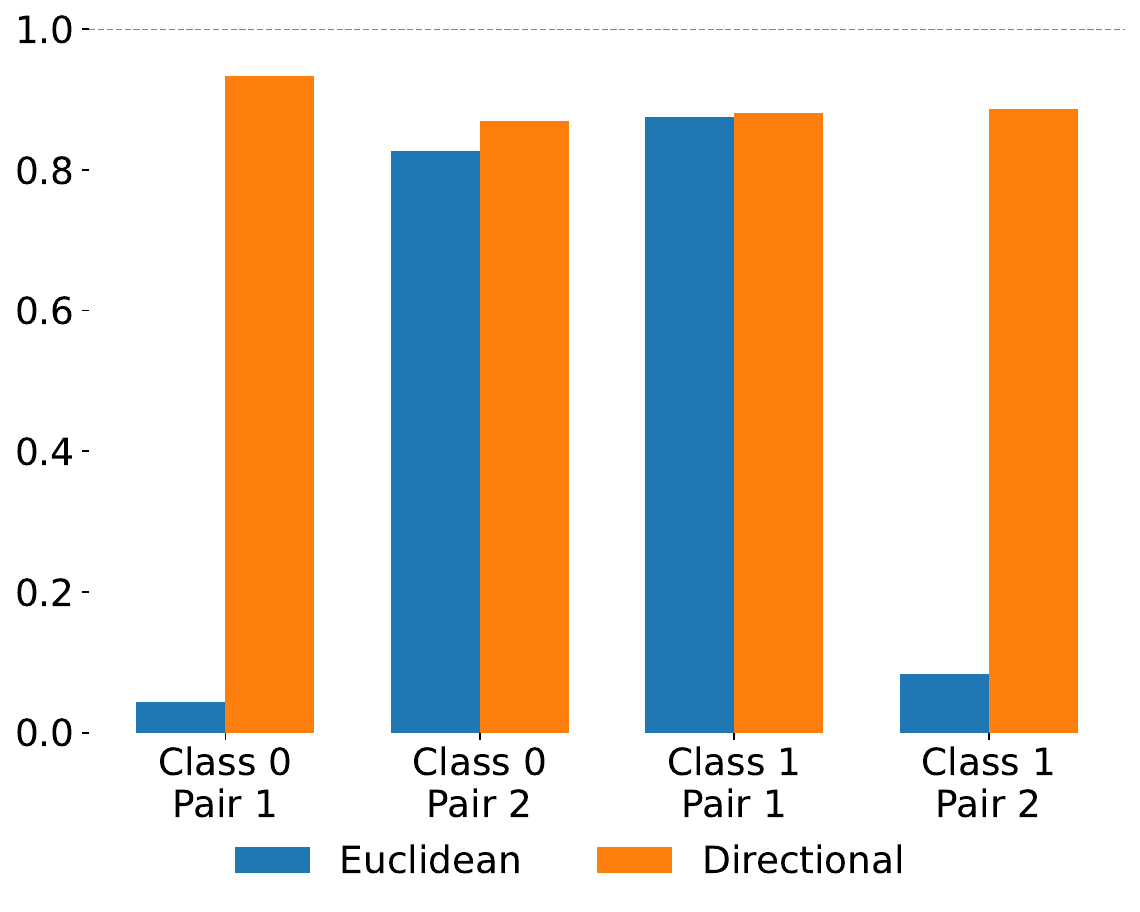}
        \caption{Summary}
    \end{subfigure}
    \hfill
    \begin{subfigure}[b]{0.59\textwidth}
        \includegraphics[width=\textwidth]{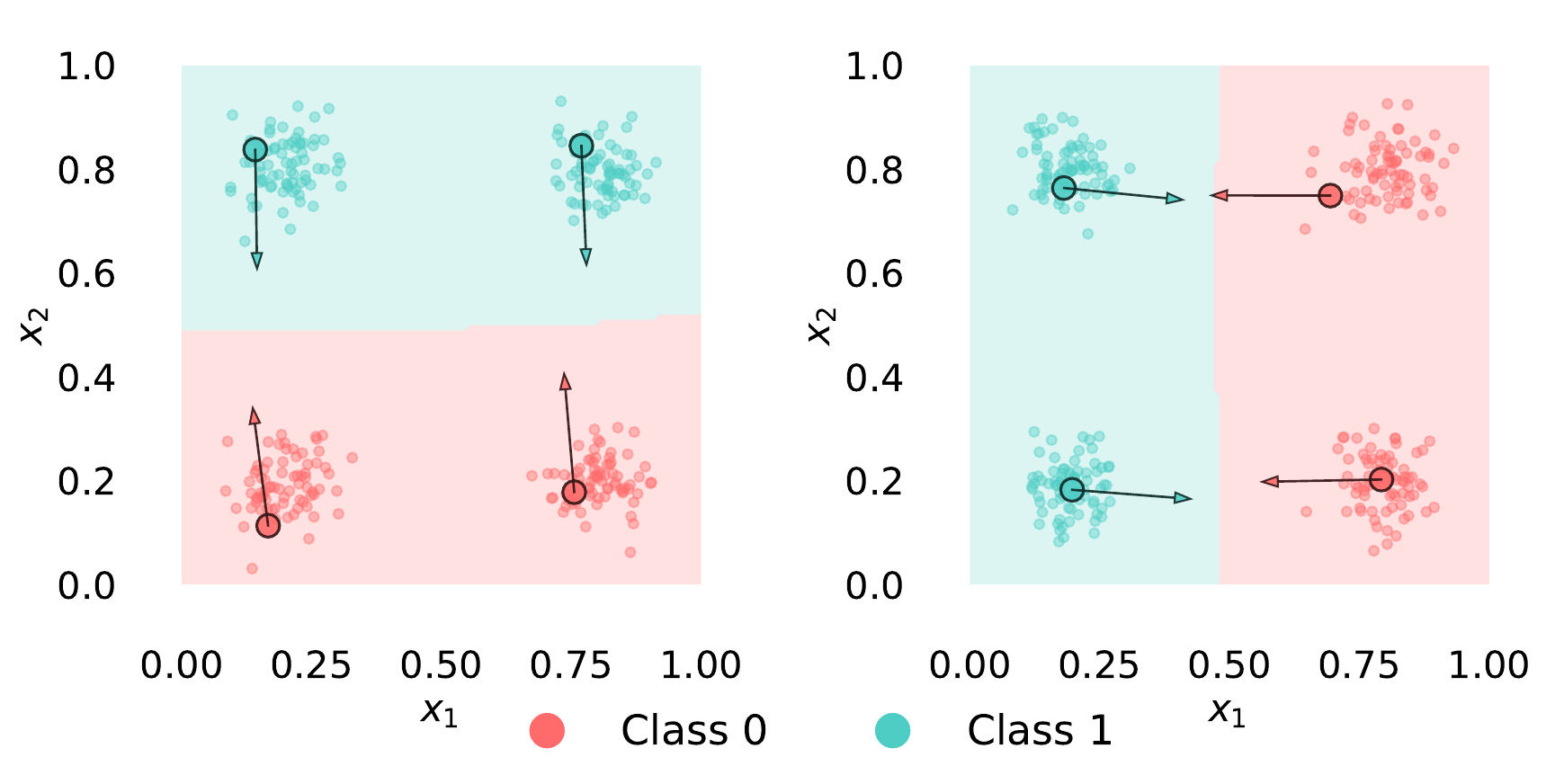}
        \caption{Feature-space visualization}
    \end{subfigure}
    \caption{Analysis of a real concept drift where two sub-concepts swap their class labels. \textbf{Global disagreement}: $d_{mae} \approx 0.5$. \textbf{Per-class feature means}: \textit{before} -- (\textbf{0:} $x_1$=0.5, $x_2$=0.2, \textbf{1:} $x_1$=0.5,$x_2$=0.8), \textit{after} -- (\textbf{0:} $x_1$=0.8,$x_2$=0.5, \textbf{1:} $x_1$=0.2,$x_2$=0.5).}
    \label{fig:case2}
\end{figure}

Next, we examine a pure real concept drift, where the underlying data distribution $P(X)$ is stationary, but the posterior probability $P(Y|X)$ is altered, leading to the sudden concept drift. Specifically, two sub-concepts exchange their class labels (Fig. \ref{fig:case2}).

\textbf{Data} The per-class feature means undergo a substantial change, correctly signaling a substantial shift in class definitions. However, this analysis alone cannot distinguish this class-swapping event from a shift of the data clusters in the feature space.

\textbf{Model} A high global disagreement ($d_{mae} \approx 0.5$) indicates a severe and abrupt drift, confirming the previous model is now obsolete. The local disagreement values of 1.0 for the two affected sub-concepts (Fig. \ref{fig:case2}(b)) successfully pinpoint the regions of instability. While this layer excels at quantifying the drift's magnitude and impact, it fails to explain its root cause.

\textbf{Explanation} The GCE analysis provides the crucial explanatory mechanism. It reveals that the centroids associated with two sub-concepts have effectively swapped their positions between classes (Fig. \ref{fig:case2}(b, d)). Moreover, drastic transformation of the CFAVs where all groups have been inverted (Fig. \ref{fig:case2}(a)), and the directional change approaches its maximum value (Fig. \ref{fig:case2}(c)) provides an interpretable signal that the model's fundamental decision logic has been rewritten.

This case highlights the synergy of our framework. The model layer quantified the severe impact, and the data layer pointed to a class-definition change. Yet, it is the explanation layer that delivers the conclusive evidence, demonstrating through the evolution of centroids and CFAVs that the drift was a re-labeling event, not a spatial data shift.

\begin{figure}[h!]
    \centering
    \begin{subfigure}[b]{0.37\textwidth}
        \includegraphics[width=\textwidth]{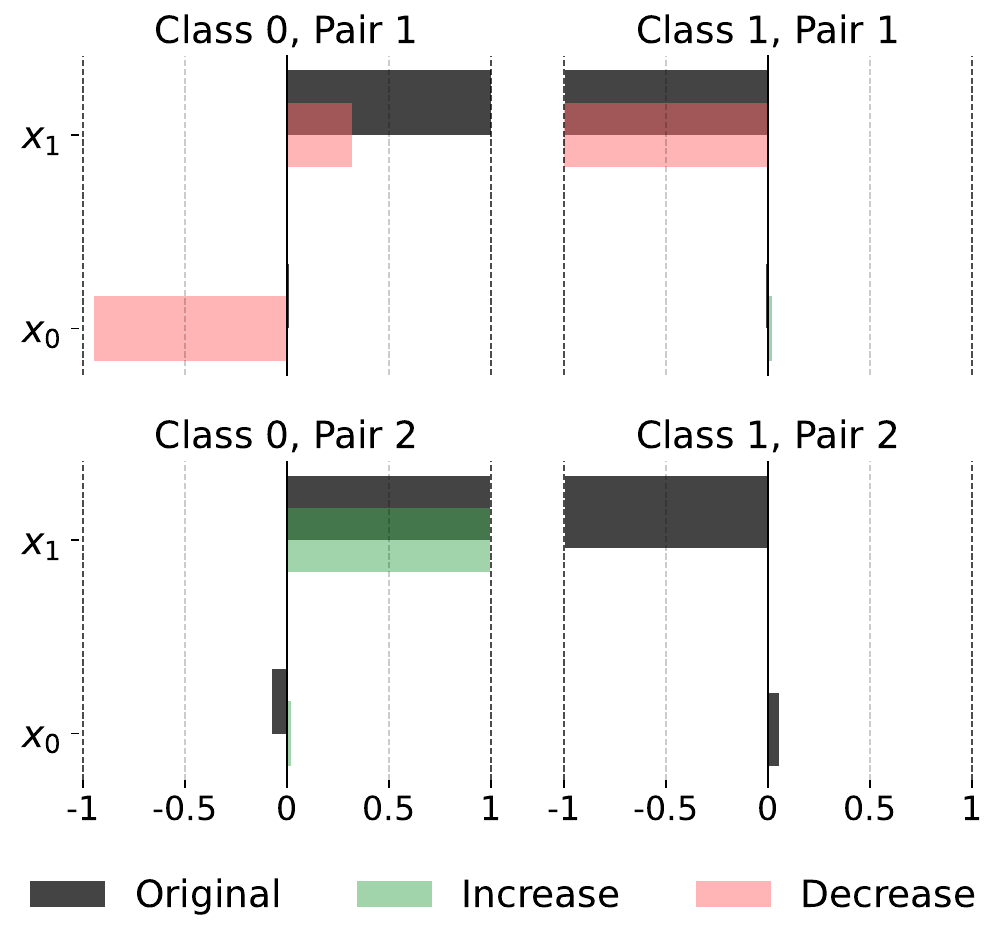}
        \caption{CFAVs}
    \end{subfigure}
    \hfill
    \begin{subfigure}[b]{0.5\textwidth}
        \includegraphics[width=\textwidth]{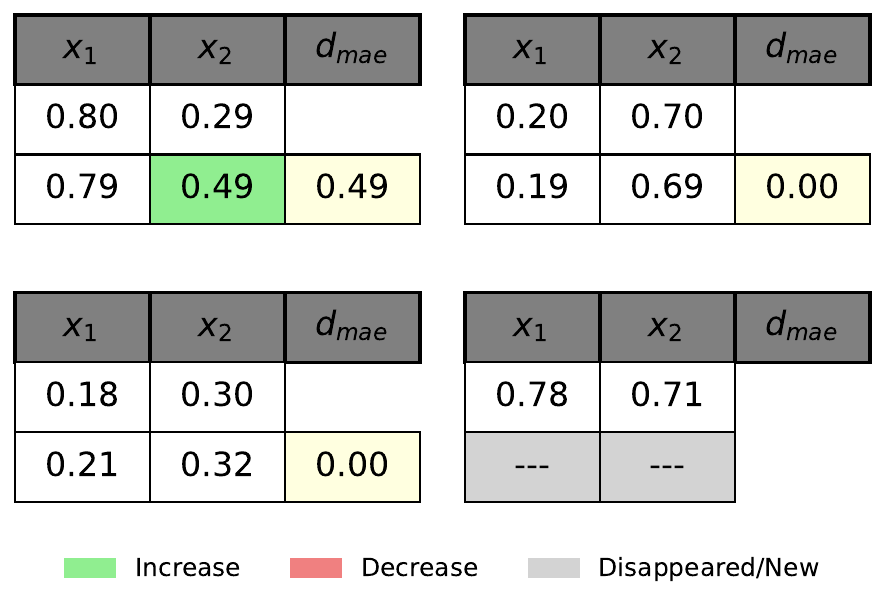}
        \caption{Centroids}
    \end{subfigure}
    \hfill
    \begin{subfigure}[b]{0.37\textwidth}
        \includegraphics[width=\textwidth]{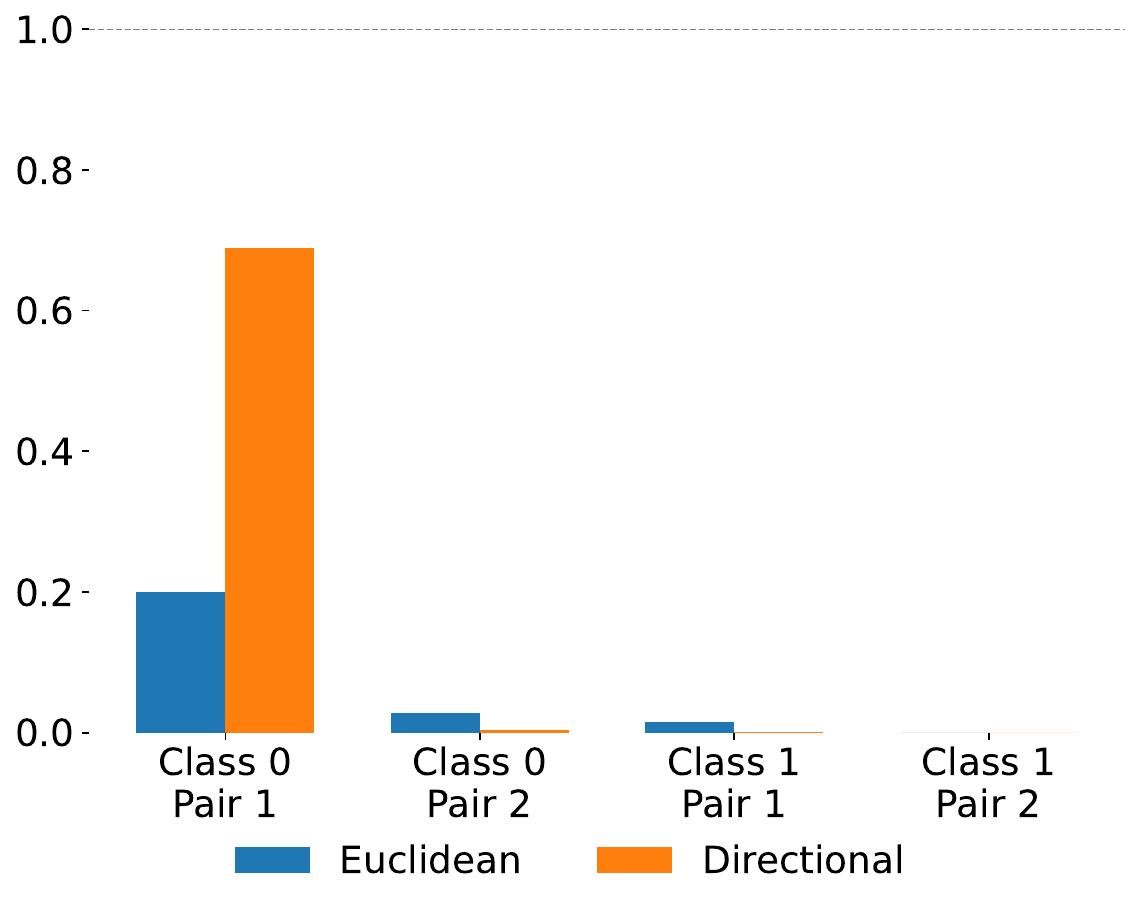}
        \caption{Summary}
    \end{subfigure}
    \hfill
    \begin{subfigure}[b]{0.59\textwidth}
        \includegraphics[width=\textwidth]{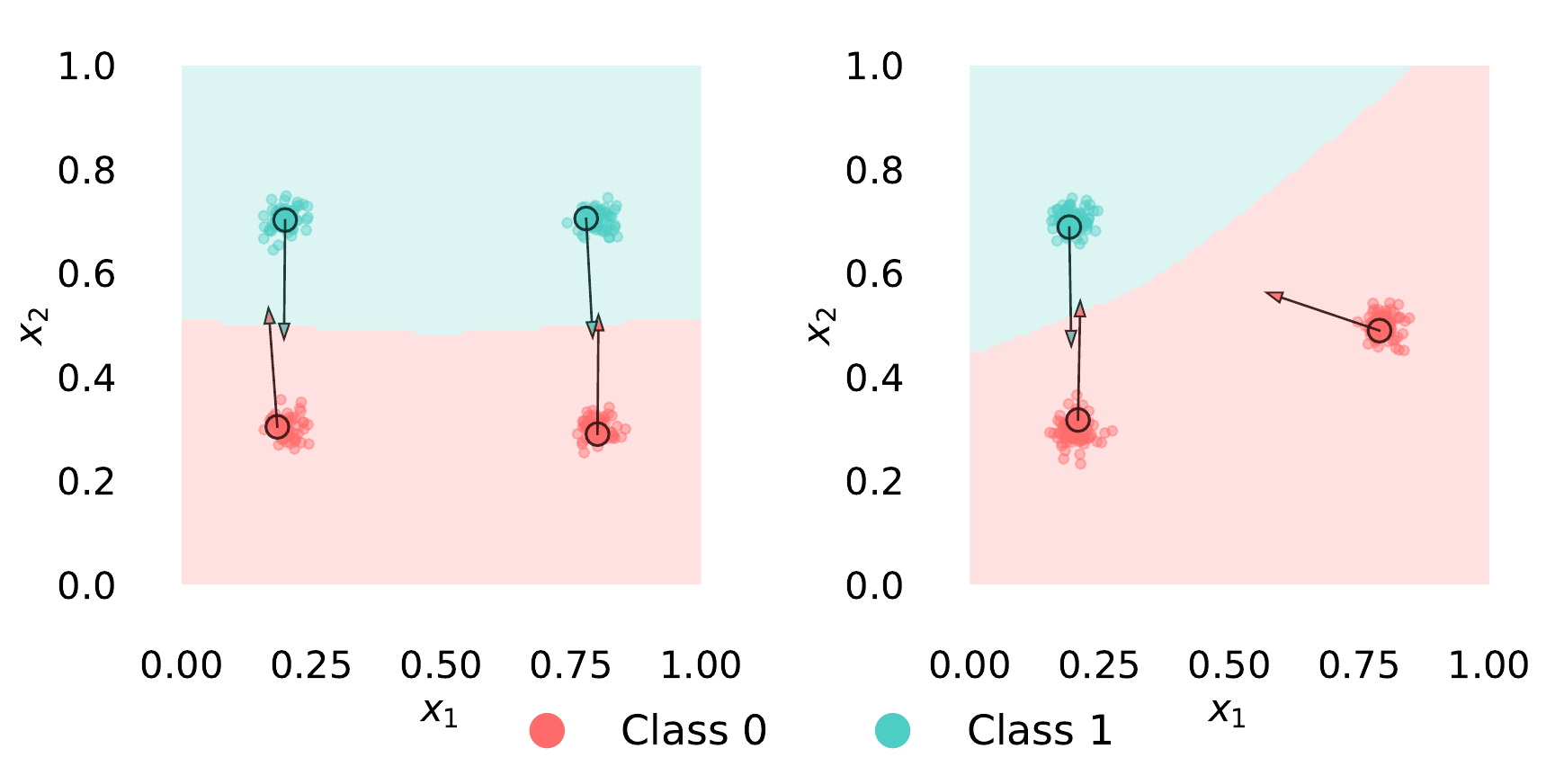}
        \caption{Feature-space visualization}
    \end{subfigure}
    \caption{Analysis of a combined drift involving both data shifts and changes in decision logic. \textbf{Global disagreement}: $d_{mae} = 0.1$. \textbf{Per-class feature means}: \textit{before} -- (\textbf{0:} $x_1$=0.5, $x_2$=0.3, \textbf{1:} $x_1$=0.5,$x_2$=0.7), \textit{after} -- (\textbf{0:} $x_1$=0.5,$x_2$=0.4, \textbf{1:} $x_1$=0.2,$x_2$=0.7).}
    \label{fig:case3}
\end{figure}

\subsection{Case 3: Combined Drift }
Finally, we analyze a more complex scenario where both $P(X)$ and $P(Y|X)$ shift simultaneously. Some data clusters move which also results in changes of a decision boundary (Fig. \ref{fig:case3}).

\textbf{Data:} The per-class means shows that the mean $x_1$ in class 1 has deceased, while $x_2$ in class 0 has slightly increased. This signals a notion of data drift but offers little clarity on its nature.  

\textbf{Model:} The global disagreement is moderate ($d_{mae} = 0.1$). On the other hand, while using information from the explanations on centroids one  can notice that the local disagreement in Fig. \ref{fig:case3}(b) is non-zero only for \textit{Class 0, Pair 1} (0.318), successfully localizing the region of highest model instability.

\textbf{Explanation:} The GCEs provide a detailed breakdown. Fig. \ref{fig:case3}(d) and \ref{fig:case3}(b) clearly show that the centroids for \textit{Class 0, Pair 1} have shifted in the feature-space. 
In parallel, the CFAV for this same pair shows a significant directional change (Fig. \ref{fig:case3}(a, c)), directly corresponding to the region of model change identified by the model layer. Meanwhile, other clusters, \textit{Class 1, Pair 1} and \textit{Class 0, Pair 1} remain stable in both location and CFAV, while \textit{Class 1, Pair 2} has disappeared entirely.

This case exemplifies the usefulness of the proposed approach. 
All layers detect a change, but only their combination provides a full diagnosis of the causes of this considered situation. 
The data layer gives a high-level alert. 
The model layer pinpoints the consequences of the drift (i.e., where and if the model's predictions became unreliable allowing for approximating the drift is not so strong). 
The explanation layer reveals the real consequences: a spatial shift of one sub-concept, a change in the decision logic for that same region, and the disappearance of another. 
This  multi-faceted explanation is aids understanding and responding to complex, real-world drifts.

\section{Discussion}
\label{conclusion}

In this paper, we introduced a novel methodology for explaining concept drift by analyzing the temporal evolution of group counterfactual explanations. By shifting focus from numerous individual explanations to a concise set of group-level recourses, our approach offers a more tractable and interpretable lens into model changes, enabling better localization of the change and an examination of its causes in the data. We operationalized GCEs within an \textit{explanation layer} that, when combined with traditional data- and model-level analyses, provides a synergistic and multi-faceted understanding of drift.

% This paragraph is strong. I've only made very slight changes for conciseness.
Furthermore, we demonstrated that this three-layer framework helps reveal underlying mechanisms of change that are difficult to identify using any single layer alone. 
By analyzing the movement of GCE centroids in feature space alongside the transformation of their associated counterfactual action vectors, our method can distinguish a spatial data shift from a fundamental re-labeling of concepts, or identify a combination of both. 
This new methodology can not only lead to a more comprehensive understanding of the reasons for concept drift by end-users, but also foster better reactions to it and improve the management of dynamic machine learning models.

\section{Limitations and Future Directions}
\label{sec:limitations}

Our analysis has demonstrated the explanatory power of evolving GCEs, primarily using synthetic datasets with well-defined sub-concept structures. This controlled setting was instrumental for validation, yet it highlights several important areas for future research. 
A primary consideration is the performance of the methodology on data lacking a clear cluster structure. 
Our current pipeline determines the number of groups, $k$, using x-means, an approach that is inherently less effective on uniform or diffuse distributions. 
Subsequently, applying GLANCE, an algorithm whose clustering metric is not well suited for such data, can lead to a suboptimal partitioning of the feature space. This underscores the need to investigate alternative GCE generation methods that can uncover relevant groupings in diverse data topologies.

Yet another future challenge includes carrying out similar studies with drifts occurring in real data streams, also affected by additional sources of data difficulties, such as class overlap, noise or outliers. 

Beyond the diagnostic power of explaining drift, the use of GCEs in data streams presents a practical use-case opportunity, which provides efficient, low-latency counterfactual recourse. 
By amortizing the cost of explanation across a small set of shared action vectors, a system could avoid the expensive task of computing a new CE for every single instance. 
However, this creates a key research challenge in developing methods to dynamically maintain these GCEs as the underlying data stream evolves.\\ 

\noindent
\textbf{Acknowledgement}: This work has been supported by the National Science Centre, Poland (Grant No. 2023/51/B/ST6/00545).

\bibliographystyle{plainnat} 
\bibliography{ref}

% \appendix
% \input{chapters/6appendix}

\end{document}